# Rigid Transformations for Stabilized Lower Dimensional Space to Support Subsurface Uncertainty Quantification and Interpretation

Ademide O. Mabadeje[a, *] and Michael J. Pyrcz[a, b]
[*]Corresponding author E-mail: ademidemabadeje@utexas.edu

[a]Hildebrand Department of Petroleum & Geosystems Engineering, The University of Texas at Austin, 200 East Dean Keeton Street, Stop C0300, Austin, Texas, 78712, USA.
[b]Department of Geological Sciences, The University of Texas at Austin, 2305 Speedway, Stop C1160, Austin, Texas, 78712, USA.

## Abstract

Subsurface datasets commonly are big data, i.e., they meet big data criteria, such as large data volume, significant feature variety, high sampling velocity, and limited data veracity. Large data volume is enhanced by the large number of necessary features derived from the imposition of various features derived from physical, engineering, and geological inputs, constraints that may invoke the curse of dimensionality. Existing dimensionality reduction (DR) methods are either linear or nonlinear; however, for subsurface datasets, nonlinear dimensionality reduction (NDR) methods are most applicable due to data complexity. Metric-multidimensional scaling (MDS) is a suitable NDR method that retains the data's intrinsic structure and could quantify uncertainty space. However, like other NDR methods, MDS is limited by its inability to achieve a stabilized unique solution of the low dimensional space (LDS) invariant to Euclidean transformations and has no extension for inclusions of out-of-sample points (OOSP). To support subsurface inferential workflows, it is imperative to transform these datasets into meaningful, stable representations of reduced dimensionality that permit OOSP without model recalculation.

    We propose using rigid transformations to obtain a unique solution of stabilized Euclidean invariant representation for LDS. First, compute a dissimilarity matrix as the MDS input using a distance metric to obtain the LDS for $N$-samples and repeat for multiple realizations. Then, select the base case and perform a rigid transformation on further realizations to obtain rotation and translation matrices that enforce Euclidean transformation invariance under ensemble expectation. The expected stabilized solution identifies anchor positions using a convex hull algorithm compared to $N + 1$ case from prior matrices to obtain a stabilized representation consisting of the OOSP. Next, the loss function and normalized stress are computed via distances between samples in the high-dimensional space and LDS to quantify and visualize distortion in a 2-D registration problem. To test our proposed workflow, a different sample size experiment is conducted for Euclidean and Manhattan distance metrics as the MDS dissimilarity matrix inputs for a synthetic dataset.

    The workflow is also demonstrated using wells from the Duvernay Formation and OOSP with different petrophysical properties typically found in unconventional reservoirs to track and understand its behavior in LDS. The results show that our method is effective for NDR methods to obtain unique, repeatable, stable representations of LDS invariant to Euclidean transformations.

In addition, we propose a distortion-based metric, stress ratio (SR), that quantifies and visualizes the uncertainty space for samples in subsurface datasets, which is helpful for model updating and inferential analysis for OOSP. Therefore, we recommend the workflow's integration as an invariant transformation mitigation unit in LDS for unique solutions to ensure repeatability and rational comparison in NDR methods for subsurface energy resource engineering big data inferential workflows, e.g, analog data selection and sensitivity analysis.

# 1. Introduction

Most data science studies of spatial systems require analyzing and processing big datasets [1–3]. For example, robust, closed-loop optimization, assisted history matching, and multi-source, multi-physics data assimilation require large volumes of data with a large variety and number of features, known as high dimensionality [4–8]. The curse of dimensionality is a well-known set of consequences intrinsic to working with high-dimensionality datasets. This includes poor sample coverage, insufficient data to estimate joint probabilities, and high prediction model variance due to sparse sampling and commonly observed multicollinearity all common in the high dimensional space [9]. In addition, many systems are only effective for relatively low dimensional data, and when the data dimensions are higher than the system's tolerance, data is inadequately processed and models are unreliable. For example, these high dimensional data pose challenges for building and training machine learning models, due to a loss of sensitivity in distance measures often essential for notions of distance or dissimilarity [10–12]. Thus, to process high dimensional systems, dimensionality reduction is crucial.

Dimensionality reduction (DR) is the transformation of high-dimensional datasets into meaningful representations of reduced dimensionality used for data modeling, compression, visualization, and extracting relevant information. Obtaining unique and comparable representations of reduced dimensionality is required when performing statistical analysis, modeling, and inferential analysis between original and reduced dimensional spaces in the subsurface. This is essential for uncertainty quantification and interpretation workflows. Examples of such applications include model error treatments in history matching via data-space inversion, identifying subsurface resource analogs, and comparing the performance of training image-based geostatistical algorithms for unconditional simulations [4, 7, 13–15]. In addition, low dimensional spaces are used to predict outcomes in other fields, for example, predicting hit dance songs in music [16].

Originally developed on the tenets of linear algebra (e.g., principal components analysis), DR methods were developed through statistics and machine learning across different disciplines, making them indispensable tools for analyzing high-dimensional data. In DR, there is an a priori assumption about the shape and dimension of a low-dimensional manifold where these high-dimensional datasets reside. A manifold is a topological space that is locally Euclidean, i.e., every point on the manifold has a neighborhood that is topologically equivalent to an open subset of Euclidean space. Thus, DR methods attempt to find low-dimensional representations of highly dimensional datasets using coordinate representations by constructing a linear or nonlinear manifold [17]. A linear manifold is a subset of a closed vector space under vector addition and scalar multiplication. Equivalently, it is defined as a subset of a vector space that can be expressed as the span of a set of vectors. An example of linear manifolds in two-dimensional spaces is lines and points, which can be thought of as lines and planes in three-dimensional spaces, and lines, planes and hyperplanes in > three-dimensional spaces while in higher $d$-dimensional spaces. Hyperplanes can describe tangent planes to a smooth hypersurface, allowing for the study of the geometry of the hypersurface in a linear context. A DR method is defined as linear if for a dataset with sample size, $n$, $d$-dimensional data points $X = [x_1, ..., x_n] \in \mathbb{R}^{d \times n}$ with a dimensionality of $r < d$. Therefore, it inherits all the algebraic properties of a vector space defined by the linear manifold. An objective function $f(\cdot)$ is optimized to yield a linear transformation, $P$, such that $P \in \mathbb{R}^{d \times n}$, resulting in deterministic low-dimensional transformed data, $X'$, where $X' = PX \in \mathbb{R}^{R \times n}$. There exist many linear dimensionality reduction methods (LDR) such as principal component analysis (PCA) [18, 19], Fisher's linear discriminant analysis [20, 21], locality preserving projections [22, 23], along with important extensions using kernel mappings [24, 25] and regularizations on a linear manifold [17, 26].

However, these LDR methods cannot adequately deal with subsurface datasets, which are complex, nonlinear, and sometimes governed by unknown geological and engineering assumptions. Thus, more flexible nonlinear manifold dimensionality reduction (NDR) methods, also known as manifold learning, are preferred. NDR methods attempt to find a nonlinear manifold using local neighborhoods, geodesic distances, or graph theories to obtain a space of reduced dimensionality by preserving the intrinsic or extrinsic structures of the data. Commonly used NDR methods in the subsurface domain include local linear embeddings (LLE) [27], multidimensional scaling (MDS) [28–30], isometric mapping (IsoMap) [31], Laplacian Eigenmaps [32], t-distributed stochastic neighbor embedding (t-SNE) [33], and diffusion maps [34]. However, Cunningham and Ghahramani [35] showed that most NDR methods could be fundamentally formulated as a more general form of MDS, thereby serving as the basis of various methods and subsequent extensions, as seen in IsoMap and t-SNE.

Multidimensional scaling originates from the work of Young and Householder [36], which examines the relationship between a set of points concerning mutual pairwise distances in Euclidean space by establishing necessary and sufficient conditions to approximate these points in a space of lower dimensionality. Subsequently, Torgerson [28] expanded upon this work by developing procedures to obtain projections from dissimilarities between sets of points in the feature space to obtain a space of reduced dimensionality. Many variations exist of MDS, including nonmetric MDS, an iterative approach to obtain the representation of reduced dimensionality in ordinal datasets, and metric MDS, methods applicable to multivariate datasets. Both variants rely on a dissimilarity metric of choice or Euclidean distance function as inputs.

Metric MDS is a method that maximizes the scatter of the projection in Eq. (1) such that the most informative mapping is obtained using pairwise dissimilarities, $d_{ij} = d_X(x_i, x_j)$, opposed to the actual sample values. The maximization of the scatter of these projections are performed iteratively using the SMACOF (Scaling by Majorizing a Complicated Function) optimization algorithm initialized with a pseudo-random number generator popularly referred to as a random state. The projected points, $x'_i$, $x'_i = \theta^T x_i$ for some $\theta \in \mathcal{O}^{d \times r}$ whilst maximizing pairwise distances, $\sum_i \sum_j \|x'_i - x'_j\|^2$, in the low dimensional space (LDS), where $\theta$ is the optimization variable. LDS is a meaningful mapping from high to low dimensional space maps similar input vectors to nearby points on a manifold. Thus, metric MDS can be formulated as a generalized eigenvalue problem resulting in matrix decomposition from optimizing the scatter objective below [35].

$$f(\theta) = \sum_{i,j} \left( d_X(x_i, x_j) - d'_{X'}(\theta x_i, \theta x_j) \right)^2, \qquad (1)$$

where $d_X(x_i, x_j)$ is the pairwise dissimilarities between points in the feature space and $d'_{X'}(\theta x_i, \theta x_j)$ is the scatter projection to be maximized in the LDS. Equation (1) above implies that $X'$ is a nonlinear mapping of $X$, which can be computed using a variety of distance metrics for the dissimilarity matrix computation, **D**, depending on the motivation, objective functions, and data input of choice [37].

However, most manifold learning algorithms work in batch mode, where all data points are used during training, thus the method is unsuitable for a sequential data stream (i.e., when a new sample is added to the data) as the entire algorithm needs to be recalculated with the original dataset augmented by the newly added samples, making it computationally expensive. This can result in intractable computations for subsurface applications such as real-time monitoring or online production optimization. For these methods, it is not possible to insert new samples in the data, i.e., out-of-sample objects (OOSP) into an LDS that represents the original objects in Euclidean space, Trosset et al. [38] formulate the problem as an out-of-sample embedding problem. Examples of such approaches are in Kouropteva et al. [39], Law et al. [40], and Bengio et al. [41], where incremental and out-of-sample extensions for LLE, IsoMap, Eigenmaps, MDS, and spectral clustering NDR methods are developed, making it applicable to sequential data streams. Other methods to embed OOSP are typically done with different neural network architectures; however, this is contingent on a large sample size for training [42].

Alternatively, Verleysen and Lee [43] propose combining information visualization from user-centric graphic objectives and lower dimensionality representations from NDR to visualize $d$-dimensional datasets. User-centric graphic objectives are based on controllability (a user-defined way to represent the data) and interaction. However, the effect of these concepts in visualizing representations from NDR methods is complicated due to problems posed in terms of sensitivity to initial conditions, rotations, reflections, translations, and affine transformations. Furthermore, these NDR methods require a predefined optimization criterion for a given dataset, which raises the need to rerun the algorithm making it computationally and time expensive when new samples are added. Although the embedding quality in NDR methods is mostly satisfactory, a shared limitation exists of providing an embedding only for the inputted dataset, with no extension for OOSP.

Lee and Verleysen [44] developed a rank-based quality criterion to quantify the embedding quality of an LDS in terms of its intrinsic and extrinsic structure preservation for different DR methods. Recall that an embedding representation of reduced dimensionality, i.e., LDS, is obtained for all NDR methods. However, due to conflicting requirements and objectives between DR methods, there is a need to identify a generalizable metric for visualization quality in NDR methods. Hence, it is imperative to investigate and answer the question of how stability alters the representation of data in LDS in order to develop effective and usable visualizations.

Stability is the property of an LDS representation to remain consistent and unchanged under small perturbations or variations in the input data, multiple model runs with the same data, or NDR algorithm parameters [45]. The notion of stability stems from concerns around visual perception, interpretability, and obtaining several equivalent representations from a single NDR method. Suppose the parameters of NDR methods are altered, even infinitesimally, then instability can occur, resulting in a very different representation that is nearly equivalent to the model loss function but distinct from the original representation. This instability type is undesirable in subsurface LDS visualization and interpretation as it makes inferential analysis inconsistent.

Computer vision, remote sensing, medical sciences, and robotics have extensive developments for understanding representations and configurations of stationary or moving objects in space. This concept is central to solving registration problems related to images. Registration is a technique that spatially aligns two or more images from different sources taken at different times and angles into a single integrated image through the identification and alignment with anchor points. For example, in the positioning of wireless sensor networks to identify the current location of sensor nodes, a known subset of pairwise distances between various sensors and a subset of anchor locations is used to infer the remaining sensors [46]. Anchors are spatial points that ensure the pose of an object will be continuously updated to maintain a fixed position and orientation in space [47]. Typically, registration is performed to overcome Euclidean or spatial transformations that are commonplace when overlaying images, videos, or datasets using either rigid or non-rigid transformations as needed [48–51].

The inability to attain unique LDS solutions invariant to translation, rotation, and reflection, i.e., Euclidean transformations, is peculiar to NDR methods. Rigid transformations, also known as Euclidean transformations, are defined as a set of mathematical operation or geometric transformation that preserves the shape and size of an object from one location to another by moving, flipping, or changing a shape to create an image from an existing preimage [49–51]. A preimage is the original figure or collection of points before a transformation(s), while the image is the final figure or reoriented set of points obtained after a transformation(s) is applied. An image is distinct from the preimage as it shows the changes made during the transformation process. For example, a reflection transforms a preimage by flipping across the line of reflection to create an image. Thus, rigid transformations are not subject to shearing and non-uniform scaling transformations to create identical preimages from an image.

Motivated by the abovementioned limitations, we propose a novel workflow for rigid transformation for stabilized lower dimensional space, MDS invariant to Euclidean transformations to improve interpretability and understanding of LDS such that appropriate inferences alongside uncertainty space visualizations can be made for OOSP. Our proposed workflow computes the input dissimilarity matrix using the Euclidean distance measure via metric MDS to obtain a representation of the LDS. Rigid transformation is performed on multiple MDS realizations for the batch model $N$ case and sequential $N + 1$ case to address stability in the LDS. This is done to ascertain Euclidean transformation invariance via identifying sample data anchor points, making the workflow applicable to OOSP. Lastly, to improve inferential analysis and quantify the embedding quality of representations in LDS, we use normalized stress to quantify the distortion between $N + 1$ and $N$ case samples inclusive of an OOSP, which serves as a diagnostic of model quality. Our proposed workflow addresses the problem of non-unique solutions obtained via MDS due to Euclidean transformations by providing a robust and generalizable framework, which creates a stable representation of the LDS that is invariant while providing an exact solution supporting the inclusion of out-of-sample-points to support new workflows. Examples include real-time model updating and data assimilation, predictive models for well performance, predictive classification of undrilled well's productivity, and prediction of hit music.

The methodology section details our proposed workflow for LDS instability control at an algorithmic level; an invariant methodology to stabilize lower dimensional representations. The result section describes the subsurface datasets in high and low dimensional spaces, governing assumptions, and associated limitations of the proposed workflow. Lastly, we explore the accuracy and validity of the proposed workflow concerning subsurface datasets of high dimensionality from wells in the Duvernay unconventional reservoir.

## 2. Methodology

Our proposed LDS stabilization workflow applies to representations obtained from all manifold dimensionality reduction methods. Greater details are included for an MDS-based approach for illustration. For the metric MDS-based example, these are the steps:

1. Standardize the predictor features, $P$, in high dimensional space with sample size, $N$, to calculate $P^s$.
2. Compute the sample dissimilarity matrix input, $D$ using an appropriate dissimilarity metric.
3. Perform metric MDS on the standardized predictor features, $P^s$, to obtain a realization of the lower dimensional space for the base case, $\mathbf{Z}_B$.
4. Repeat step (3) over a large number of realizations, $L_k$, using different random states for the initialization(s) of SMACOF in metric MDS.
5. Ascertain each realization of the LDS, $\mathbf{Z}_k$, obtained over $L_k$ are on the same scale; otherwise, normalize it.
6. Compute the rigid transformation method for each $\mathbf{Z}_k$ and $\mathbf{Z}_B$ such that the optimal rotation and translation matrices for each representation denoted by $\widehat{\mathbf{R}}_k$, and $\widehat{\mathbf{T}}_k$ is obtained respectively.
7. Apply $\widehat{\mathbf{R}}_k$, and $\widehat{\mathbf{T}}_k$ on $\mathbf{Z}_k$, to compute the stabilized solution, $\mathbf{S}_k$ for each realization.
8. Compare $\mathbf{S}_k$ to $\mathbf{Z}_B$ and assess the model performance using normalized stress, $\sigma_N$ as a measure of goodness and distortion measure. If $\mathbf{S}_k$ exhibits significant distortion, proceed to step (5) and repeat subsequent steps.
9. Determine the convex hull polygon of the expectation of all stabilized solutions, $\mathbf{S}_E$, and identify the anchor points.
10. Include an OOSP within 95% confidence interval for each predictor feature, and obtain the LDS representation for a single realization, $\mathbf{Z}_{OOSP}$, with a sample size of $N + 1$ and repeat for all $L_k$ realizations.
11. Determine the convex hull polygon and identify the anchor points from the $N + 1$ sample size inclusive of the OOSP in the LDS.

12. Perform rigid transformation on the two anchor point sets to obtain the optimal $\widehat{\mathbf{R}}_{OOSP}$, and $\widehat{\mathbf{T}}_{OOSP}$ matrices, which is applied on all datapoints in the $N + 1$ sample case in the LDS to compute the stabilized solution, $\mathbf{S}_{OOSP}$.
13. Compare $\mathbf{S}_{OOSP}$ to $\mathbf{S}_E$ and check the solution with the normalized stress, $\sigma_{OOSP}$, and determine the stress ratio (SR).

The first step of our proposed method for LDS stabilization is to standardize the predictor features of interest, $P$, $\forall\, m = 1, \ldots, M$, and $M = \{\, m \mid m \in \mathbb{N}\}$, each separately to standard normal distribution to obtain its respective standardized predictor, such that $P^S \sim N(0,1)$ with a sample size of $N$. This is performed to assist with outliers and provide better interpretations and inferences, e.g., remove sensitivity to choice of feature's units, etc. Next, the dissimilarity matrix, $\mathbf{D}$, with elements denoted by $d(i,j)$, is computed using the Euclidean distance as its dissimilarity metric in Eq. (2), which is employed on $P^S$ over the available $m$ predictors in the feature space for all $i, j$ data pairs where $n$ is the total number of data pairs.

$$d(i,j) = \sqrt{\sum_{m=1}^{M}(p_{i,m} - p_{j,m})^2} \quad \forall\, m = 1, \ldots, M;\ i, j \in \{1, \ldots, n\}, \qquad (2)$$

$$d(i,j) \colon \mathbb{R}^m \to \mathbb{R}^m.$$

Metric MDS is computed using $\mathbf{D}$ as input to obtain the LDS representation of the base case, $\mathbf{Z}_B$, with a reduced dimensionality $\mathbb{R}^2$. Afterward, metric MDS is recomputed over a large enough number of $L_k\ \forall\, k = 1, \ldots, K$, and $K = \{\, k \mid k \in \mathbb{N}\}$ realizations using a series of different random states to change the initialization of the SMACOF solutions found via metric MDS to obtain the $\mathbf{Z}_k$. If the LDS representation of each realization is not on the same scale, perform any normalization technique of choice, e.g., univariate standardization of each MDS projection for each $\mathbf{Z}_k$ realization.

Based on the work of Yang et al. [52], which showed that the LDS representation obtained from MDS is unique except for Euclidean transformation, i.e., translation, rotation, or reflection of the entire solution space; therefore, the overall shape or structure of an MDS representation can be transformed without changing its essential properties or the relationship between data points. Recall that rigid transformation (RT) is a method that reconstructs a given set of points ($\geq 3$) on a plane between two objects, using Euclidean transformations to ensure object similarity is preserved shown in Fig. (1) [49–51]. There exist different rigid transformation implementations in literature; however, based on the work of Eggert et al. [49], the singular value decomposition algorithm (SVD) is selected as the preferred choice for RT with $N \leq 1000$ applicable to tabular datasets. Hence, the proposed method utilizes the SVD approach developed by Arun et al. [53], applicable to three-dimensional datasets, which is modified and made applicable to two-dimensional datasets as well. For a more detailed explanation of rigid transformations, see the works of Eggert et al. [49] and Miyakoshi [50].

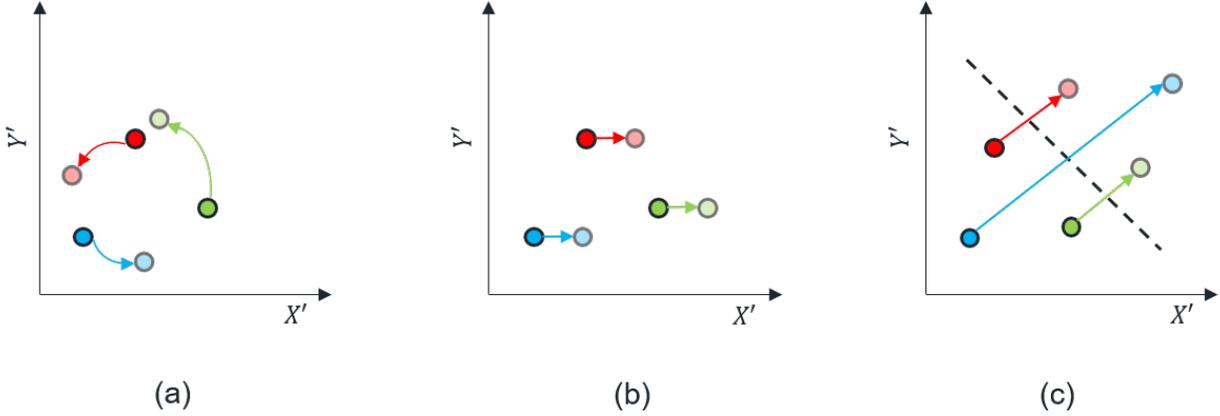

(a) (b) (c)

**Fig. 1** A schematic showing the three possible Euclidean transformation types: (a) rotation, (b) translation, and (c) reflection. The preimage is represented as dark-colored circles, and the image is represented by the colored circles with lesser intensity, which are transformed to match the preimage using rigid transformations.

The sixth step in the workflow is to compute an SVD-based rigid transformation between each $\mathbf{Z}_k$ (image) and $\mathbf{Z}_B$ (preimage) in the LDS based on the assumption that the two set points are related by Eq. (3).

$$\mathbf{Z}_B = \mathbf{R}_k \mathbf{Z}_k + \mathbf{T}_k + \mathbf{V}_k , \qquad (3)$$

where $\mathbf{R}_k$, $\mathbf{T}_k$, and $\mathbf{V}_k$ are the rotation or flip, translation, and noise matrix, respectively. The centroids in the LDS representation are determined for the image and preimage using Eqs. (4) and (5), where $\bar{c}_B$ and $\bar{c}_k$ represents the respective centroids of the preimage and image(s) in the LDS MDS projections for each realization performed.

$$\bar{c}_B = \frac{1}{N}\sum_{i=1}^{N} \mathbf{Z}_B, \qquad (4)$$

$$\bar{c}_k = \frac{1}{N}\sum_{i=1}^{N} \mathbf{Z}_k, \qquad (5)$$

$$\bar{c}_B , \bar{c}_k : \mathbb{R}^2 \to \mathbb{R}^2.$$

Next, mean centering is performed $\mathbf{Z}_B$ and $\mathbf{Z}_k$ to ensure $\bar{c}_B$ and $\bar{c}_k$ are at the origin to remove the translation component such that the least squares error (LSE) criterion in Eq. (6) is minimized $\forall\ L_k$ realizations.

$$\mathrm{LSE} = \sum_{i=1}^{N}\|\mathbf{Z}_B^* - \mathbf{R}_k \mathbf{Z}_k^*\|^2, \qquad (6)$$

where $\mathbf{Z}_B^*$, is the mean-centered vector of the preimage, $\mathbf{Z}_k^*$ is the mean-centered vector of the image(s) $\forall\ L_k$ realizations, $\mathbf{R}_k$ is the rotation matrix $\forall\ L_k$ realizations. From SVD, every real square matrix, $\mathbf{H}$ can be expressed as a product of three matrices, $\mathbf{H} = \mathbf{USV}^T$, where $\mathbf{U}$ and $\mathbf{V}$ are orthonormal matrices of the left and right singular vectors respectively, and $\mathbf{S}$ is a diagonal matrix.

Thus, it can be proven that performing SVD on an accumulating matrix, $H_k = Z_B^* Z_k^{*T} \; \forall \; L_k$ realizations, there is an optimal rotation matrix, $\widehat{R}_k$, for each realization that maximizes the desired trace, which is calculated via Eq. (7). Note that Eq. (7) suffices only when the determinant of $\widehat{R}_k = +1$, if the determinant of $\widehat{R}_k < -1$, then a reflection transformation has occurred and the $\widehat{R}_k$ is found using Eq. (8) where $V'_k$ is the same as the $V_k$ with one or more columns having negative signs.

$$\widehat{R}_k = V_k U_k^T, \tag{7}$$

$$\widehat{R}_k = V'_k U_k^T, \tag{8}$$

Meanwhile, the optimal translation matrix, $\widehat{T}_k$ is that which aligns the centroids of the preimage, $c_B$, with the rotated centroids of the image, $c_k$, described in Eq. (9).

$$\widehat{T}_k = c_B - \widehat{R}_k c_k. \tag{9}$$

In steps seven through eight, the optimal $\widehat{R}_k$ and $\widehat{T}_k$ obtained is applied on $Z_k$ to obtain the stabilized solution, $S_k$, such that the least square error obtained during RT is minimized and the normalized stress, $\sigma_N$, $\forall \; L_k$ realizations is minimized too in Eq. (10) [54] for the $N$-sample case. See Table A.1 in Appendix A for more details on the normalized stress criteria and interpretability.

$$\sigma_N = \sqrt{\frac{\sum_{i<j}\left(d_{i,j} - d_{i,j}^{Z_k}\right)^2}{\sum_{i<j}(d_{i,j})^2}}, \tag{10}$$

where $d_{i,j}$ is the element-wise computation of the original dissimilarities between the data points in the feature space, and $d_{i,j}^{Z_k}$ is the element-wise computation of the matrix of distances in the LDS space for each $Z_k$ for all realizations. This ensures that every $S_k$ is mathematically invariant and approximately equal to the preimage, $Z_B$ with minimal distortion. According to our derived Theorem A.1 (Proof in Appendix A), the ensemble expectation $\forall \; S_k$ gives the stabilized solution, $S_E$. Then, the Quickhull algorithm by Barber et al. [55] is implemented to find a convex hull polygon in the LDS of reduced dimensionality, $\mathbb{R}^2$. This convex hull polygon is formed by connecting the outermost points, i.e., anchor points, in the LDS denoted as a mathematical set $\{A_n\}$, where $A_n$ is the location vector of the convex hull polygon's vertices points for the $N$-sample case.

Afterward, an OOSP with parameter values within the 95% confidence interval for each predictor feature is added to the existing dataset, yielding an augmented dataset with a sample size of $N + 1$. Metric MDS is performed using a Euclidean dissimilarity metric to compute $D$ and an LDS representation, $Z_{OOSP}^k$, is obtained for each $L_k$ realization. The convex hull polygon of the $N + 1$ points in the LDS is determined, and the subsequent set of anchor points is determined, $A_{OOSP}$, where $A_{OOSP}$ is the location vector of the convex hull polygon's vertices points for the $N + 1$ sample case. Next, rigid transformation is performed on the point sets $A_n$ (preimage) and $A_{OOSP}$ (image) to obtain the optimal rotation, $\widehat{R}_{OOSP}$, and translational, $\widehat{T}_{OOSP}$, matrices that enforce mathematical similarity between the preimage and image such that the shape and size of the image is Euclidean invariant. The now stabilized solution, $S_{OOSP}$, is obtained in the LDS, then compared to $S_E$ from the $N$-sample case and the normalized stress, $\sigma_{OOSP}$ is obtained as a measure of distortion between the original dissimilarities in the feature space and the distances in the projected LDS for the $N + 1$ sample case for all realizations shown in Eq. (11) [54].

$$\sigma_{OOSP} = \sqrt{\frac{\sum_{i<j}\left(d_{i,j} - d_{i,j}^{Z_{OOSP}}\right)^2}{\sum_{i<j}(d_{i,j})^2}}, \tag{11}$$

where $d_{i,j}$ is the element-wise computation of the original dissimilarities between the data points, and $d_{i,j}^{Z_{OOSP}}$ is the element-wise computation of the matrix of distances in the projected space for $\mathbf{Z}_{OOSP} \ \forall \ L_k$ realizations.

Lastly, we extended the work of de Leeuw and Stoop [56] on raw stress normalization to develop a novel metric, stress ratio (SR) that compares and quantifies distortions between the batch model $N$ and sequential $N + 1$ model cases in Eq. (12). This is done to ascertain a precise mathematical comparison for stress obtained from LDS configurations in the $N$ and $N + 1$ sample cases are comparable and invariant under Euclidean transformations.

$$SR = \frac{\sigma_{OOSP}}{\sigma_N}, \tag{12}$$

where $\sigma_{OOSP}$ represents the stress in the sequential $N + 1$ case and $\sigma_N$ corresponds the stress in the batch model $N$ case. This metric provides a robust measure of distortion that is independent of the data's dimensionality. When $SR \sim 1$, it indicates that both the $N$ sample and $N + 1$ sample cases exhibit approximately the same level of data distortion. This outcome is typically expected as the sample size, $N$, increases because the effect of the OOSP on distortion diminishes. However, when $SR > 1$, it signifies that the stress in the $N + 1$ sample case is higher compared to the $N$ sample case. This scenario is more likely to occur when $N$ is small, as the effect of the OOSP becomes significant relative to the total sample size.

## 3. Results and Discussion

The proposed workflow is demonstrated with two datasets: 1) a synthetic dataset to show the efficacy, robustness, and generalizability of the methodology, and 2) a real dataset to show one of its many applications in the subsurface.

### 3.1. Workflow Demonstration

Our proposed workflow is demonstrated with synthetic subsurface dataset of 200 unconventional, natural gas wells with realistic ranges and preservation of general multivariate relationships between petrophysical and geological properties is utilized from GeoDataSets respository [57]. The dataset consists of six predictor features, porosity (%), permeability (mD), acoustic impedance (kg/m$^2$s $\times$ 10$^6$), brittleness ratio (%), total organic carbon (%), vitrinite reflectance (%), and a response feature of 3-month average initial gas production per day (MCFPD). Feature ranking is performed, and the top predictor features of importance are porosity (%), acoustic impedance (kg/m²s × $10^6$), and total organic carbon (%). The response feature, denoted as NIP, is binned using the thresholds of [0, 2500, 5000, 7500, 10000] MCFPD to create a categorical response feature with four levels of well productivity in the dataset, i.e., low, medium (med), high, and very high (vhigh), respectively. A subset of the dataset comprising 30 samples is selected to represent the $N$-sample case. This sample size is chosen to provide a manageable and illustrative demonstration of the workflow's functionality.

In the $N$ sample case, the three predictors are individually transformed to a standard normal distribution, and the input for metric MDS, the dissimilarity matrix, is computed using Euclidean distance as its dissimilarity metric. Next, the base case realization is obtained, and 100 realizations, each with different random SMACOF initialization states, are run for the $N$-sample case of 30 samples via metric MDS. Four realizations are shown to highlight the variation of MDS projections in the lower dimensional space (LDS) due to Euclidean transformations, as shown in Fig. (2). To ensure consistency, the base case for the preimage of interest is selected as the realization with a random state of 1825. This choice is maintained throughout our analysis for both the $N$ and $N+1$ sample cases.

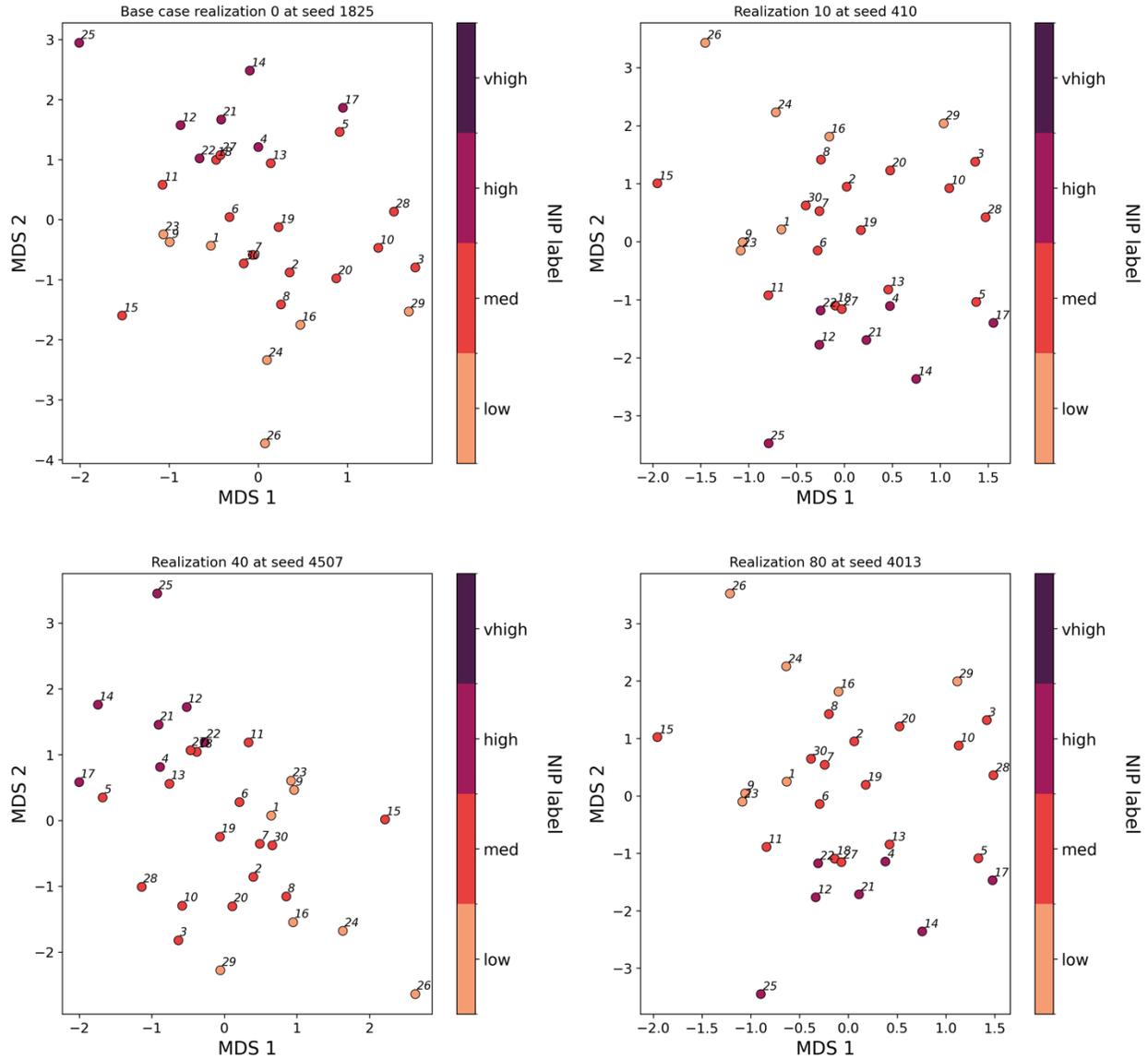

**Fig. 2** A subplot showing the MDS projections for the preimage (upper left) case and three example realizations that is color coded based on the categorical levels of normalized initial production in the *N*-sample case.

Rigid transformation between the preimage (i.e, base case) and each of the 100 realizations is applied to reorient the respective image, consistent with the preimage. This results in stabilized solutions that are invariant to Euclidean transformations, as shown in Fig. (3). At a glance, Fig. (3) demonstrates that the sample points in the example realizations are accurately registered to match the preimage, exhibiting minimal or no apparent distortions.

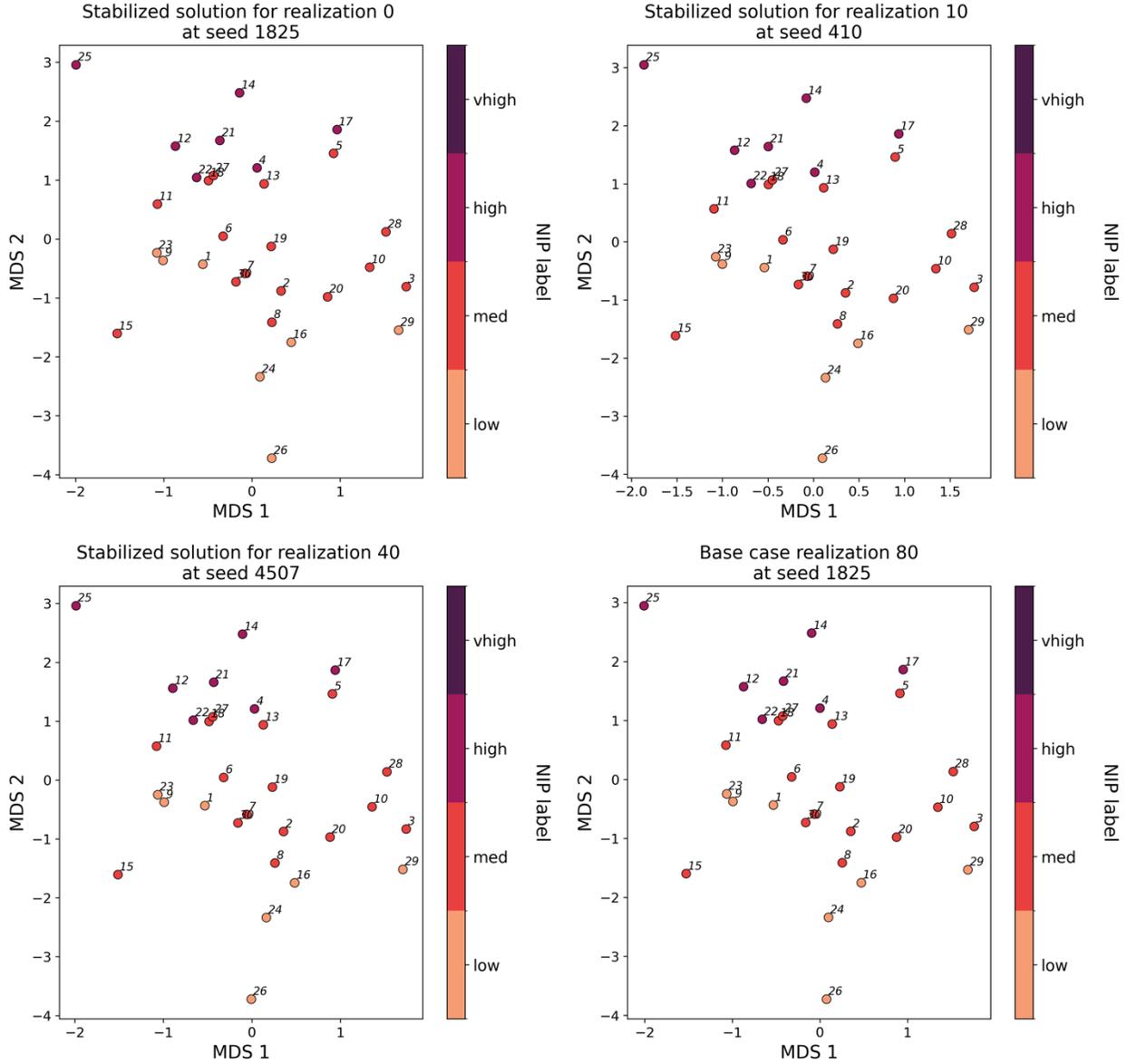

**Fig. 3** A subplot showing the MDS projections for the base case and the now stabilized three example realizations are color-coded based on the categorical levels of normalized initial production in the *N*-sample case.

To validate the stability of the rigid transformed for all 100 realizations, Fig. (4) presents the visualized stabilized solutions accompanied by the ensemble expectation derived from all 100 stabilized solutions. This comprehensive depiction captures the complete range of potential distortions, demonstrating stabilization of the lower dimensional space. In Fig. (4), it is evident that regardless of the random state used for SMACOF initialization in metric MDS, we consistently obtain approximately the same stabilized when comparing the ensemble expectation of stabilized solutions to the base case. Consequently, the ensemble expectation can serve as ground truth for all metric MDS projections in the LDS.

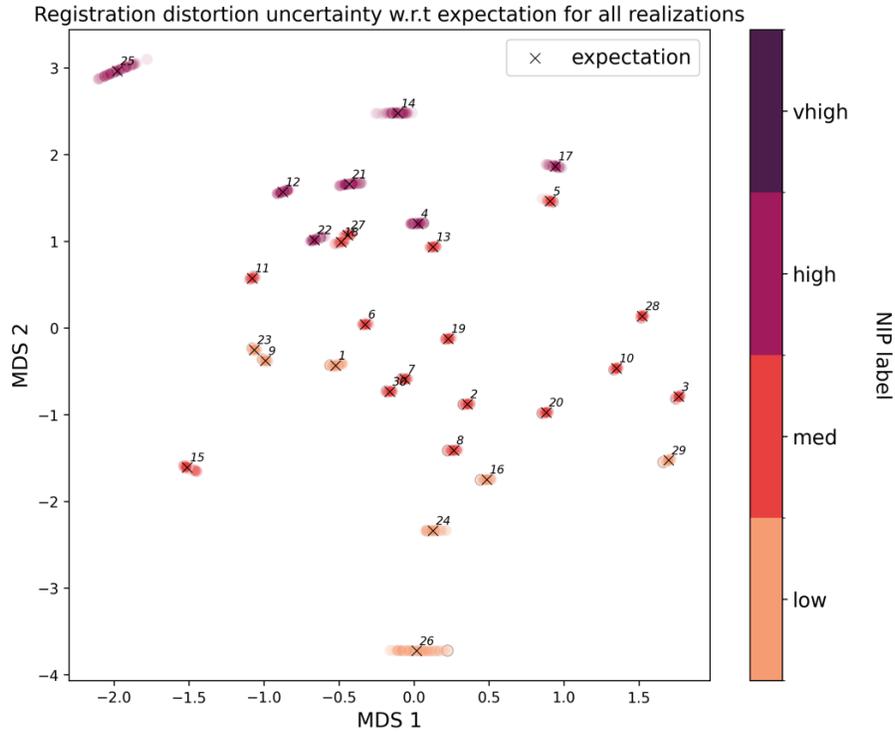

**Fig. 4** Ensemble expectation of stabilized solutions to visualize the uncertainty of distortions across all realizations. The stabilized solutions are color-coded based on the categorical levels of normalized initial production in the *N*-sample case.

Next, the ensemble expectation of the stabilized solutions in the LDS is compared to the base case on a distributional basis in Fig. (5). We observe that the kernel density estimations of the probability distribution function (PDF) for the MDS projections (MDS 1 and MDS 2) are approximately equal for both the base case and ensemble expectation.

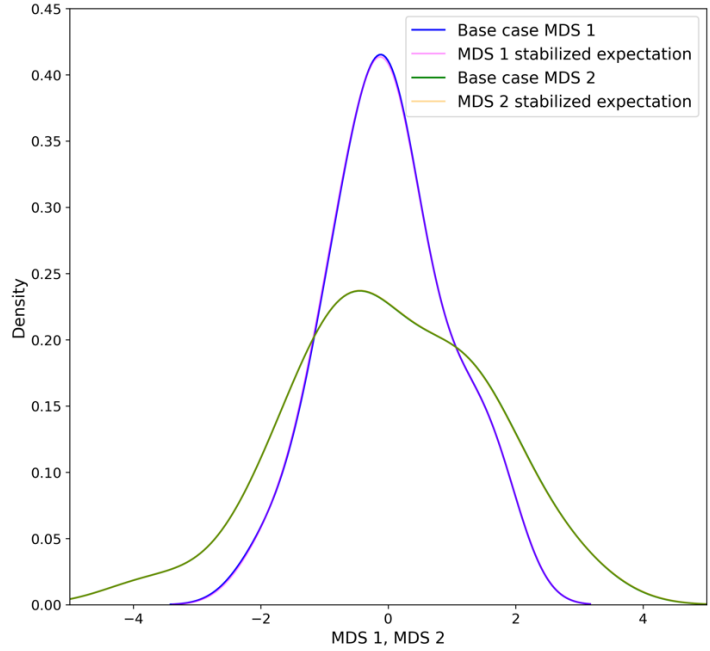

**Fig. 5** Comparison of kernel density estimation for metric MDS projections between the stabilized ensemble expectation and base case of the $N$-sample case. The x-axis is MDS projections ranges in the LDS, while the y-axis is the non-parametric PDF estimation. The overlapping projections indicates their approximate equality.

To assess the accuracy of the model, we compare the original pairwise distances in the feature space with the projected pairwise distance of the ensemble expectation in the lower dimensional space for all samples using Euclidean distances in Fig. (6). Then, a normalized stress of 0.0622 is obtained. Additionally, the convex hull polygon of the ensemble expectation of stabilized solutions in the $N$-sample case is determined, and the corresponding anchor set is identified.

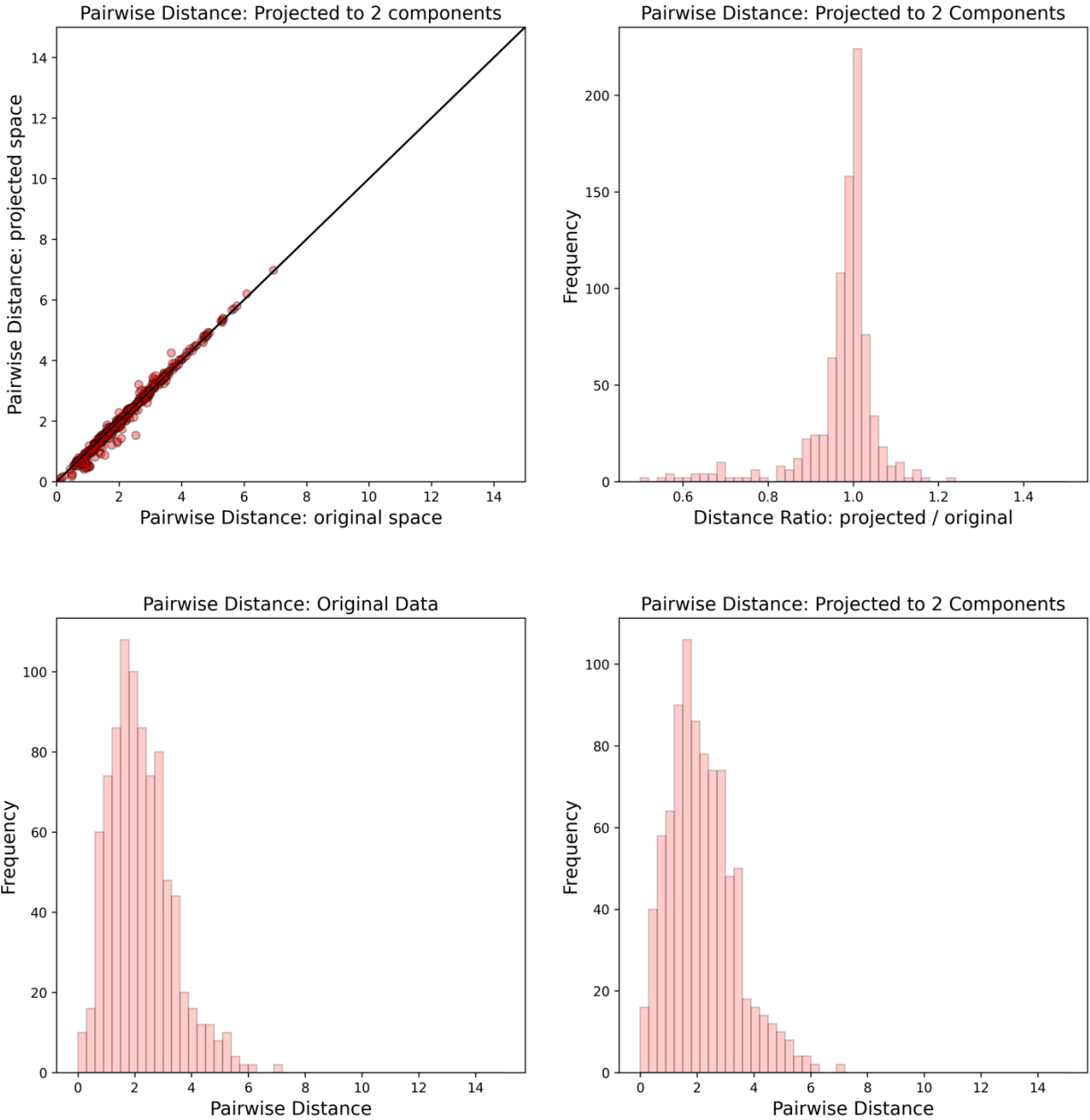

**Fig. 6** Visual model checks between the ensemble expectation of all stabilized solutions in LDS and pairwise distances in the feature space for the *N*-sample case. This comparison illustrates the preservation of pairwise distances between samples as indicated by the 1:1 line in the top left subplot, indicating the model is accurate.

Afterwards, the marginal distributions for each predictor feature are estimated, and an out-of-sample point (OOSP) is randomly generated within a 95% confidence interval for each predictor, to avoid outlier predictions. This additional sample is then added to the existing dataset to create the $N + 1$ sample case for all 100 realizations. Similar to the $N$-sample case, the workflow is repeated for all realizations using the curated $N + 1$ sample case, which includes the OOSP. By comparing the original pairwise distances in the feature space with the projected pairwise distance of the ensemble expectation in the lower dimensional space, calculated using Euclidean distances for the $N + 1$ sample case, a normalized stress of 0.0618 is obtained. The convex hull polygon of the ensemble expectation of stabilized solutions is determined for the $N + 1$ sample case, along with identifying the corresponding anchor set. Figure (7) shows the individually stabilized ensemble expectations obtained for the $N$ and $N + 1$ sample cases, with their respective anchor sets connected by a red dashed line within the identified convex hull polygons.

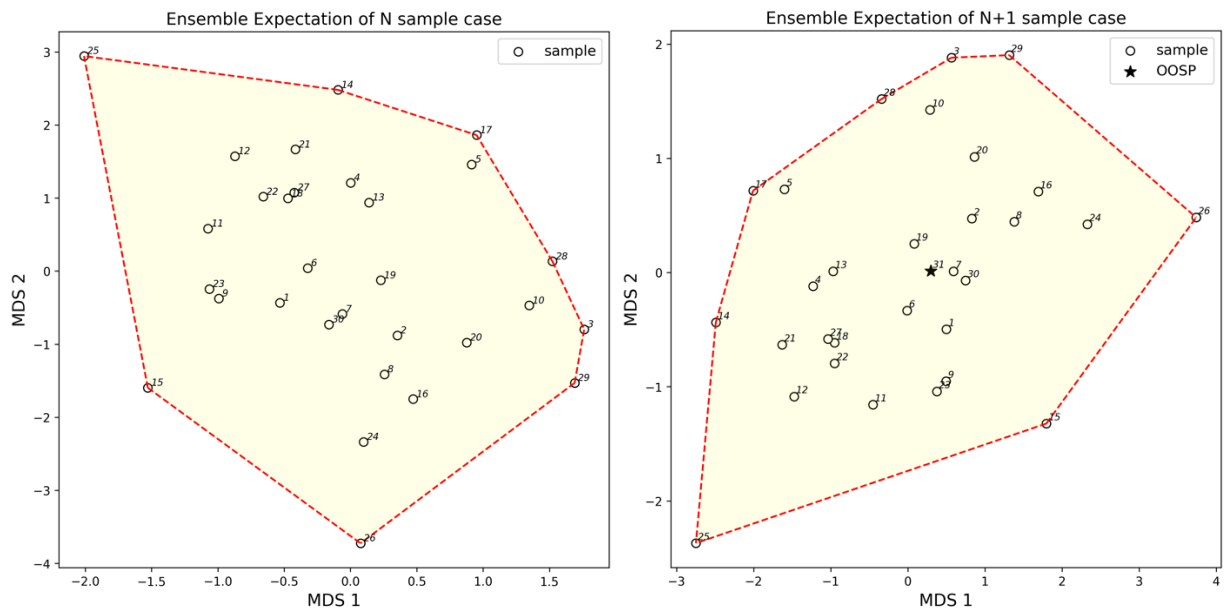

**Fig. 7** Stabilized ensemble expectation for all 100 realizations in the LDS. Left: $N$-sample case. Right: $N + 1$ sample case. Note that the ensemble expectation of the sequential model ($N + 1$ sample case) has not been corrected to align the batch model ($N$-sample case) via rigid transformations in both cases yet.

We see that although the convex hull polygons for either sample case appear different, the anchor sets consist of the same anchor points. For this example, with a sample size of 30, the anchor sets in both $N$ and $N + 1$ sample cases not only have the exact anchor points, but the vector lengths are the same, making it a relatively easier registration problem to solve using rigid transformations. However, rigid transformations cannot be readily applied when there is a mismatch in both anchor sets. Instead, the common anchor points within both sets are identified, and a rigid transformation is applied to these common anchor points. The rotation and translation matrix obtained from this transformation is then applied to the image (i.e., $N + 1$ sample case) to ensure accurate registration. This process ensures that the stabilized ensemble expectation of the image matches the ensemble expectation of the preimage (i.e., $N$-sample case).

Now, rigid transformation is performed on these two anchor sets, where the image (i.e., $N + 1$ sample case) is aligned to the preimage (i.e., $N$-sample case) to obtain a stabilized anchor solution, as shown in Fig. (8).

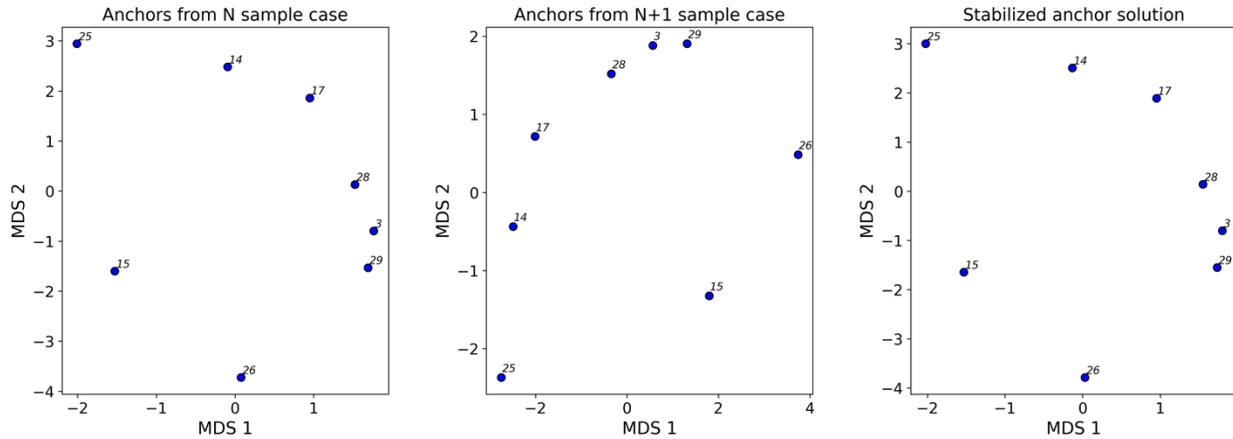

**Fig. 8** Stabilization of anchor sets from ensemble expectations. Left: $N$-sample case. Center: $N + 1$ sample case. Right: Stabilized anchor solution for $N + 1$ sample case.

By applying the rotation and translation matrices obtained through the stabilized anchor solution for the $N + 1$ sample case to all data points within the stabilized ensemble expectation of the $N + 1$ sample case, a comparison is made between the stabilized solution from the $N + 1$ sample case and the stabilized ensemble expectation of the $N$-sample case in Fig. (9). Ocularly, we observe a high level of accuracy in the obtained visualizations from these comparisons, regardless of the OOSP added.

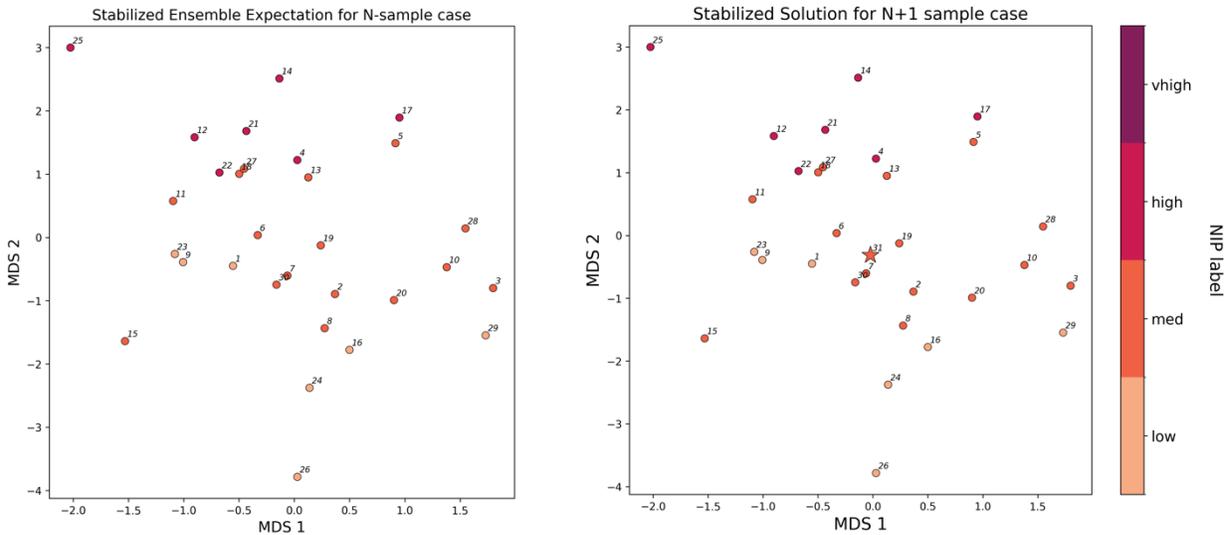

**Fig. 9** Comparison of stabilization: Color-coded representation of categorical levels of normalized initial production. Left: Stabilized ensemble expectation for $N$-sample case with original samples as circles. Right: Stabilized solution for $N + 1$ sample case with OOSP denoted by a star.

In Fig. (10), we observe that the stabilized solution for $N + 1$ sample case aligns closely with the stabilized ensemble expectation of the $N$-sample case, as demonstrated by the model accuracy visualization. This visual compares the original pairwise distances in the feature space to the projected pairwise distance of the stabilized solution for $N + 1$ sample case in the lower dimensional space, utilizing Euclidean distances for all samples. Notably, a stress ratio of 0.99 is obtained, indicating minimal distortion in both sample cases.

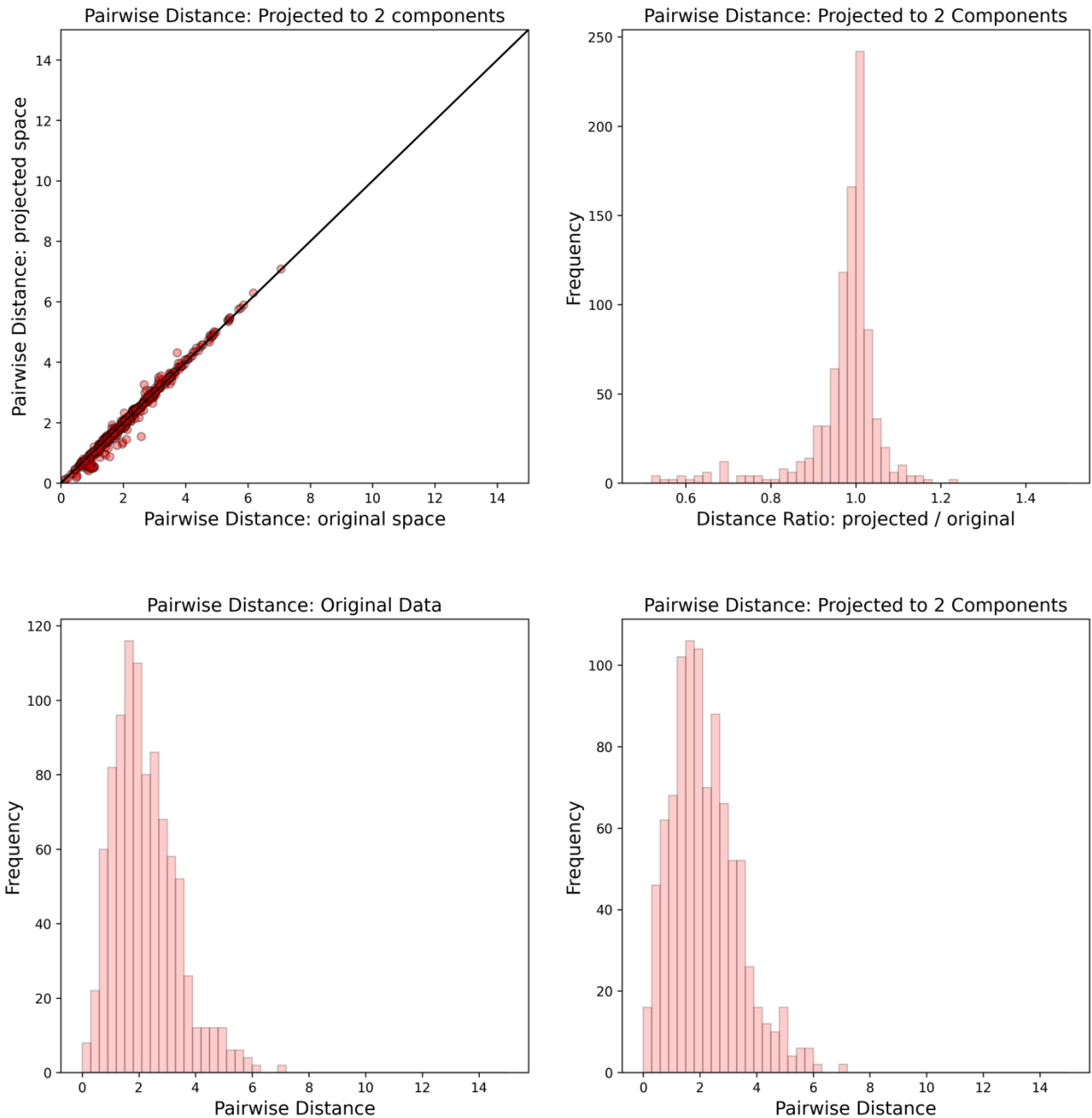

**Fig. 10** Visual model checks between the stabilized solution for the $N + 1$ sample case in the LDS and pairwise distances in the feature space. This comparison illustrates the preservation of pairwise distances between samples as indicated by the 1:1 line in the top left subplot, indicating the model is accurate.

An experiment is conducted using different sample sizes ($N = \{4, 5, \ldots, 99, 100\}$) employing Euclidean and Manhattan distance metrics as MDS dissimilarity matrix inputs to test the performance and robustness of the proposed workflow. Figures (11) and (12) show that irrespective of the dissimilarity matrix used, the normalized stress obtained for both $N$ and $N + 1$ sample cases remain relatively consistent, even when the sample size varies.

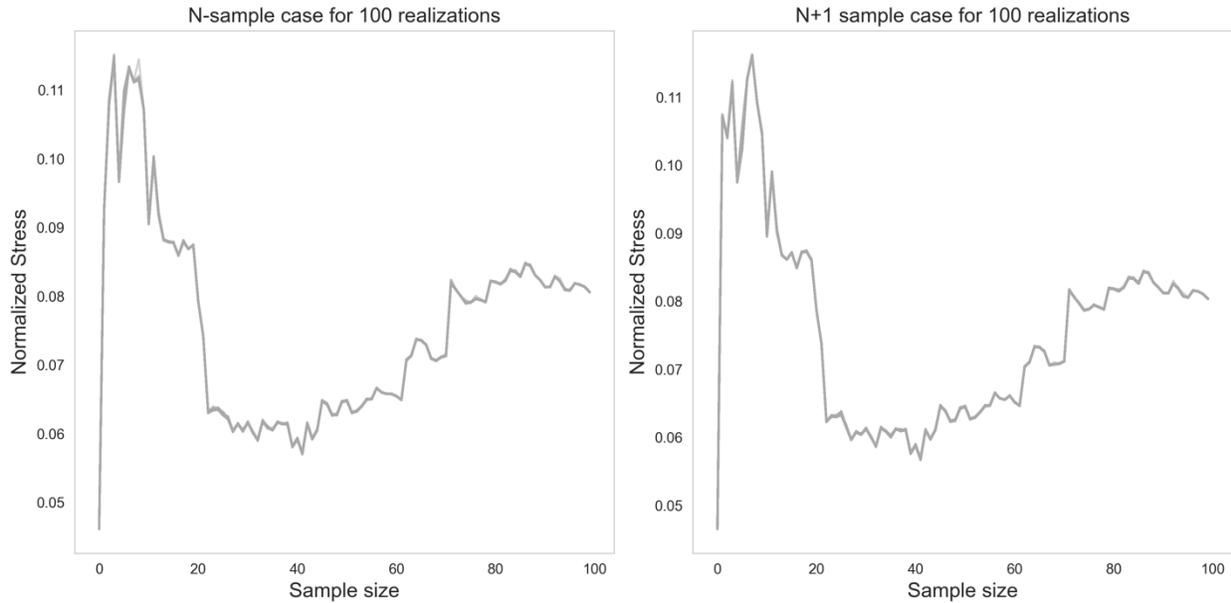

**Fig. (11)** Normalized stress levels for varying sample sizes in each run with 100 realizations, wherein each realization possesses a unique random state via a Euclidean-based dissimilarity matrix. Left: Stabilized ensemble expectation in $N$-sample case. Right: Stabilized solution in $N + 1$ sample case with a randomly generated OOSP in each experiment. Darker lines indicate each realization converges to a similar value.

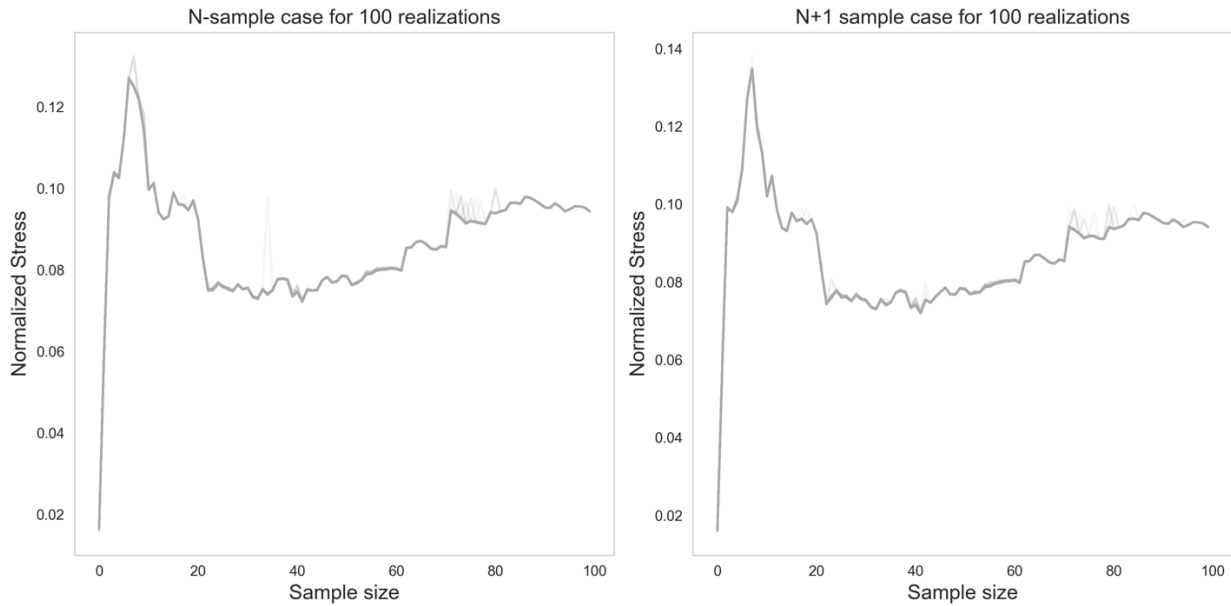

**Fig. (12)** Normalized stress levels for varying sample sizes in each run with 100 realizations, wherein each realization possesses a unique random state via a Manhattan-based dissimilarity matrix. Left: Stabilized ensemble expectation in $N$-sample case. Right: Stabilized solution in $N + 1$ sample case with a randomly generated OOSP in each experiment. Darker lines indicate each realization converges to a similar value.

Figure 13 shows that SR has high variance over small sample sizes because adding an OOSP is a more significant change to the dataset than in higher sample sizes, for both Euclidean and Manhattan regardless of the dissimilarity metric used. Also, SR converges to 1.0 with increasing $N$, indicating perfect stabilization between $N$ and $N+1$ cases.

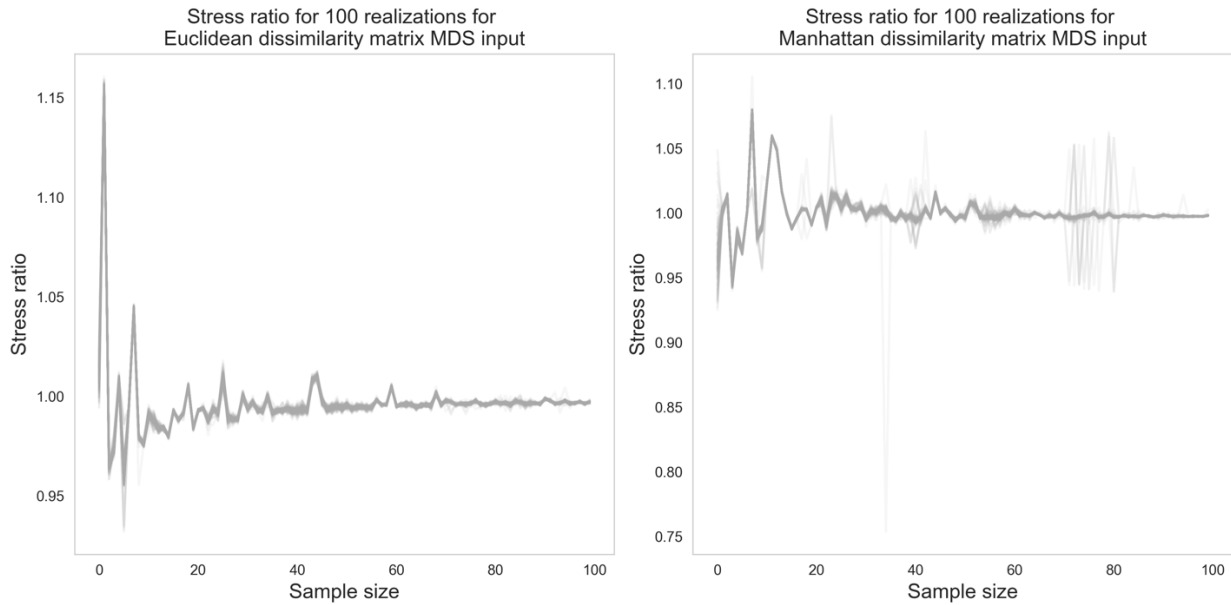

**Fig. (13)** Stress ratio for varying sample sizes in each run with 100 realizations, wherein each realization possesses a unique random state. Left: Euclidean-based dissimilarity matrix. Right: Manhattan-based dissimilarity matrix. Darker lines indicate each realization converges to a similar value.

### 3.2. Case Study with a Subsurface Dataset

Our proposed workflow is applied to a subsurface dataset of 55 unconventional shale wells consisting of petrophysical and geological predictor features from the South Kaybob field in the Duvernay formation, Canada. The sedimentary deposition of the Duvernay formation occurred during the late Devonian-Frasnian period in a sub-equatorial epicontinental seaway that coincides with the maximum transgression of the late Devonian Sea into the Western Canadian craton. The geological characteristics of the South Kaybob field reveal distinct compositional variations compared to other areas within the Duvernay. Specifically, the South Kaybob field exhibits a lower abundance of silica-rich shale and a higher clay and carbonate content than other fields. These compositional differences are attributed to the influence of the shale basin's location in the Eastern Duvernay formation. The predictor features considered are porosity (fraction), thickness (km), pore pressure (MPa), oil saturation (fraction), and a numerical response feature of cumulative gas produced (MMscf) over a northeastern location vector. The response feature is binned using the thresholds of [0, 0.5, 1.5, 2.0, 2.9] MMscf to create a categorical response feature with four levels of well productivity in the dataset, i.e., low, medium (med), high, and very high (vhigh), respectively.

The four predictor features are individually transformed to a standard normal distribution. The dissimilarity matrix, serving as the input for metric MDS, is computed using Mahalanobis distance as the dissimilarity metric. Subsequently, the base case realization is obtained, and 100 realizations, each with unique random SMACOF initialization states, are performed for the $N$-sample case via metric MDS, which is subject to Euclidean transformations variations. Rigid transformation is performed, resulting in the stabilized ensemble expectation of all realizations in the $N$-sample case, alongside its anchor sets at $\boldsymbol{\sigma_N} = 0.1943$. Then, an OOSP is randomly generated within a 95% confidence interval for each predictor and added to the $N$-sample case, creating the $N + 1$ sample case for all realizations. Like the $N$-sample case, the workflow is repeated for all realizations using the curated $N + 1$ sample case, including the OOSP. The ensemble expectation of the stabilized solutions is obtained, and the corresponding anchor set is identified at $\boldsymbol{\sigma_{OOSP}} = 0.1937$. Rigid transformation is applied on the anchor sets from the ensemble expectation for the $N$ and $N + 1$ sample cases shown in Fig. (14). Note that the normalized stress for both sample cases is ~19%; Hence, it indicates fair goodness of fit, which may be attributed to highly dimensional nature of the dataset, and reduced sample coverage.

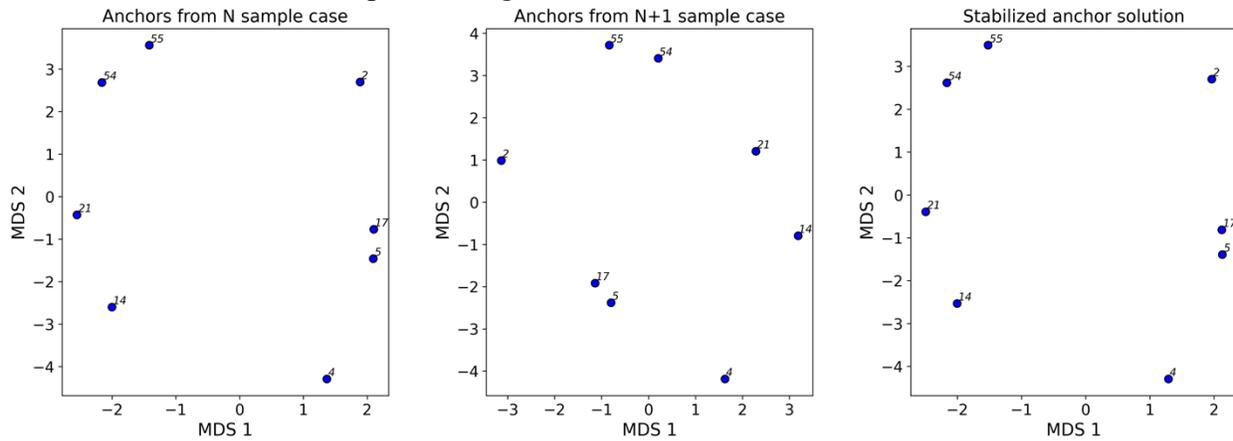

**Fig. (14)** Stabilization of anchor sets from ensemble expectations. Left: $N$-sample case. Center: $N + 1$ sample case. Right: Stabilized anchor solution for $N + 1$ sample case.

Figure (15) compares the stabilized ensemble solution for the $N$-sample case and the resulting stabilized solution for the $N + 1$ sample case. A high level of accuracy results over these cases, irrespective of the OOSP added and a $SR = 1.0$ for both cases.

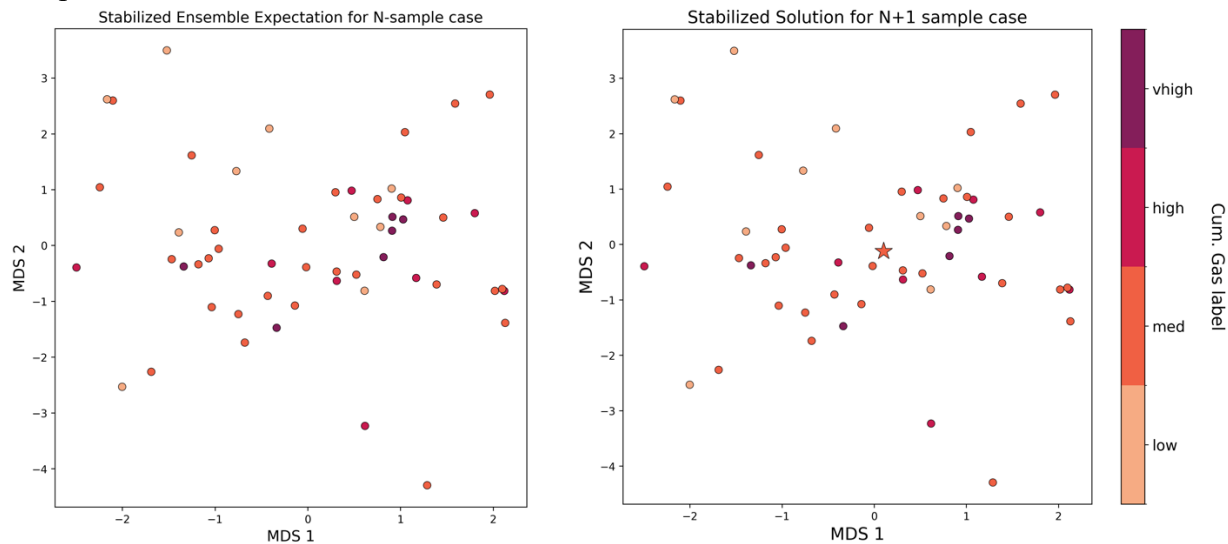

**Fig. (15)** Comparison of stabilization: Color coded representation of categorical levels of cumulative gas production. Left: Stabilized ensemble expectation for $N$-sample case with original samples as circles. Right: Stabilized solution for $N + 1$ sample case with OOSP denoted by a star.

One of the potential applications of our proposed workflow is its ability to map the trajectory of an OOSP in the lower dimensional space as predictor feature values are changed, with applicability in the characterization of oil and gas resources. This can be utilized to monitor the OOSP predictively and identify predictor combinations that facilitate the movement of the OOSP towards desired productivity regions in LDS. By selectively adjusting single or multiple predictors at a time in the LDS, the OOSP can be effectively shifted, and the methodology can be extended to cover the entire combinatorial lower dimensional space. We show this by varying the magnitude of porosity for the single-predictor scenario and the magnitudes of porosity and pore pressure for the multi-predictor scenario. The variation is performed using a multiplier, $\zeta$, that decreases and increases the predictor value of the existing OOSP parameters $\zeta \in \{0.9, 0.95, 1.0, 1.05, 1.1\}$ over the kriged gas production surface [58].

Figures (16) and (17) demonstrate the implementation of our workflow, allowing us to track the OOSP in the LDS based on varying single and multi-predictors. In both figures, the Euclidean space for the OOSP is unchanging since a fixed location vector is used. Figure (16) demonstrates a gradual reduction in the OOSP's porosity, leading to a transition from its initial well type in high-producing regions to a centralized well type in a more productive location. Further decreasing the OOSP's porosity causes the centered well type to adopt a sparse well configuration in a region of moderate production within the LDS. Similarly, gradually increasing the OOSP's porosity causes the initial well type to a shift to a visually distinct cluster and eventually adopt a sparse well configuration in regions characterized by lower production within the LDS.

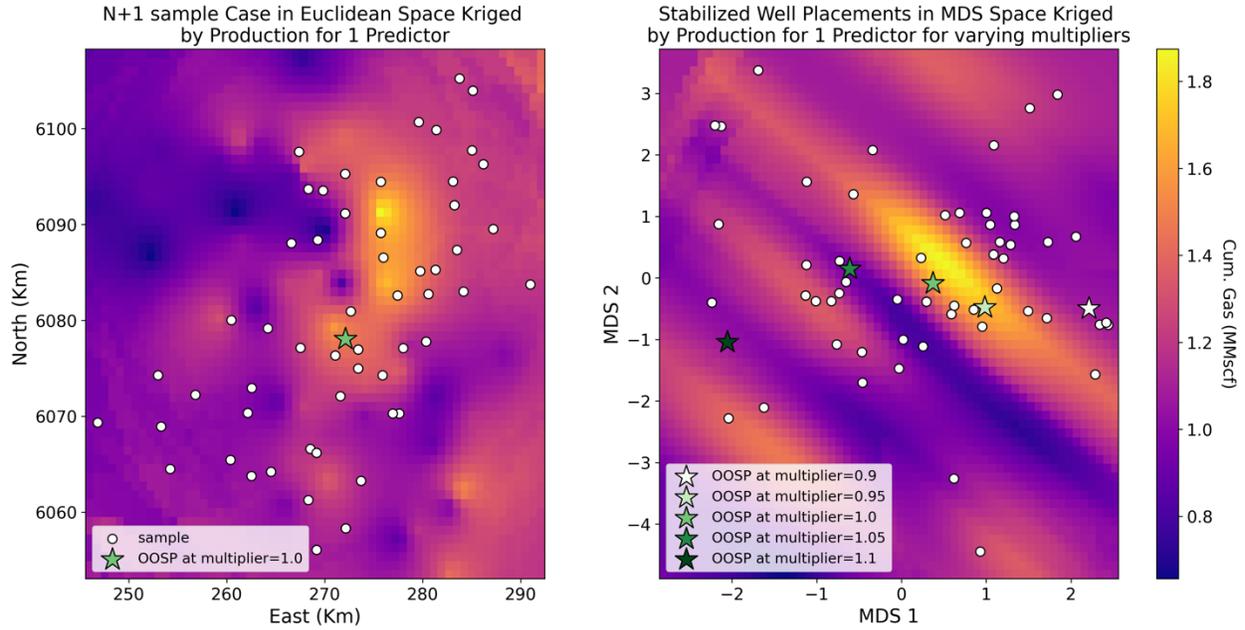

**Fig. (16)** Kriged cumulative gas production maps with data samples for single-predictor, porosity for different multipliers. Left: Euclidean space. Right: Trajectory of OOSP in LDS; as porosity increases, the OOSP moves westwards in the LDS.

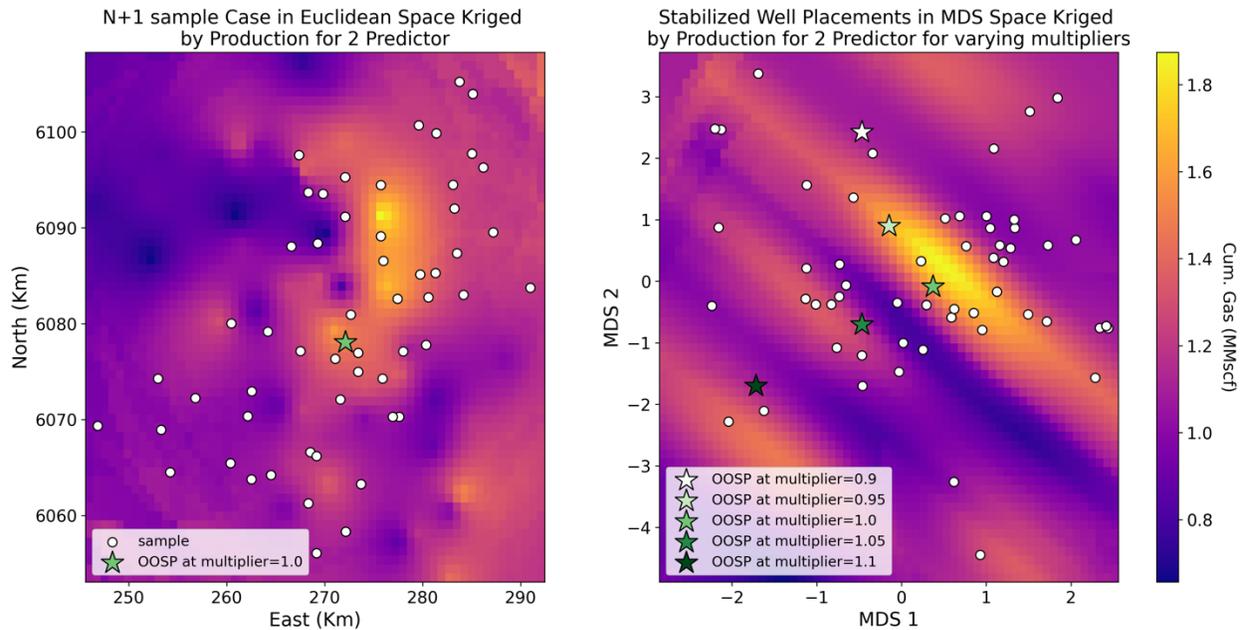

**Fig. (17)** Kriged cumulative gas production maps with data samples for multi-predictors, porosity and pore pressure for different multipliers. Top: Euclidean space. Bottom: Trajectory of OOSP in LDS; as the predictors increases, the OOSP moves southwards with a slight deviation towards the west in the LDS

Figure (17) shows a gradual reduction in the OOSP's porosity and pore pressure leading to a transition from its initial well type in high-producing regions to a somewhat sparse well type in a more productive location. Further decreasing the OOSP's porosity and pore pressure causes the well type to adopt a sparse well configuration in regions characterized by lower production within the LDS. Similarly, gradually increasing the OOSP's porosity and pore pressure causes the initial well type to remain centralized in a region of moderate production, while a further OOSP porosity and pore pressure increase results in a sparse well configuration in regions characterized by moderate production within the LDS. Therefore, decreasing both the single-predictor (i.e., porosity) and the multi-predictor (i.e., porosity and pore pressure) moves the OOSP positively in MDS 1 and is relatively constant in MDS 2 for the former, while the OOSP moves positively then remains relatively constant in MDS 1 and positively in MDS 2 for the latter. The general trend observed when increasing the multiplier from 0.9 to 1.1 shows that for both the single-predictor (i.e., porosity) and the multi-predictor (i.e., porosity and pore pressure) scenarios, the OOSP moves negatively in MDS 1, and has negative movements in both MDS 1 and 2 in the LDS, respectively.

## 4. Conclusions

Our proposed workflow overcomes limitations in metric MDS, enabling the inclusion of out-of-sample points and facilitating tracking between Euclidean or feature space and lower-dimensional spaces, while ensuring invariance to Euclidean transformation. By stabilizing the LDS, we achieve a robust framework that allows for the extraction of summary statistics, analytical workflows, and visualizations, making LDS suitable to a wide array of subsurface and beyond applications. This comprehensive approach enhances the understanding, interpretability, and applicability of MDS and LDS, paving the way for advancements in dimensionality reduction and data analysis. The methodology can also be extended to other manifold non-linear dimensionality reduction methods and parallelized for handling larger datasets, with opportunities for further research. The contributions made herein, supported by experimental evaluations and theoretical justifications, demonstrates its significant impact not only in the subsurface but also other fields.

# Appendix A

Based on *Definition 5.8.5* in Lal and Pati [59], which states a map $T: \mathbb{R}^n \to \mathbb{R}^n$ is said to be a rigid motion if $\|T(\mathbf{x}) - T(\mathbf{y})\| = \|\mathbf{x} - \mathbf{y}\|, \forall \mathbf{x}, \mathbf{y} \in \mathbb{R}^n$, where $T$ is the rigid transformation operator such that when applied on set points, $\mathbf{x}, \mathbf{y}$ each with dim $N \times 3$, leads to an invariant matrix. Thus, if a rigid transformation operator is performed on an MDS realization that preserves pairwise dissimilarity in the lower dimensional space, the same properties from the prior definition are implied and an invariant dissimilarity matrix is obtained. The mathematical proof by rigor for rigid transformations is in Sorkine-Hornung et al. [60]. Hence, we propose Theorem A.1.

**Theorem A.1**: Let $\mathbf{S}_E$ be the ensemble expectation of rigid transformed solutions for multiple realizations in the workflow known as stabilized solutions, i.e., $\mathbf{S}_E = \mathbb{E}[T(\mathbf{Z}_k)]$, in a lower-dimensional space. If an out-of-sample-point (OOSP) that falls within a 95% confidence interval for each predictor features of interest, $P \ \forall \ m = 1, \ldots, M$, and $M = \{m \mid m \in \mathbb{N}\}$ is added, then the stabilized solution obtained by applying rigid transformation on the representation obtained from the low dimensional space, $\mathbf{Z}_{OOSP}$, is the same as the ensemble expectation of the stabilized solution for the $N$-sample case, which is Euclidean transformation invariant i.e., $\mathbf{S}_E \approx \mathbf{S}_{OOSP}$.

**Proof**: Suppose adding an OOSP within the 95% confidence interval causes the stabilized solution obtained by rigid transformation on $\mathbf{Z}_{OOSP}$ to differ from $\mathbf{S}_E$. Let $T$ be the rigid transformation such that $T: \mathbb{R}^n \to \mathbb{R}^n$. This implies there is no applicable rigid transformation that ensures $\|T(\mathbf{x}) - T(\mathbf{y})\| = \|\mathbf{x} - \mathbf{y}\|, \forall \mathbf{x}, \mathbf{y} \in \mathbb{R}^n$ when $\dim\{\mathbf{x}\} \neq \dim\{\mathbf{y}\}$. This contradicts the definition of a rigid transformation as stated in Lal and Pati [59] and Sorkine-Hornung et al. [60].

Therefore, assuming the shape and size of the sets of anchor points from the $N$-sample and $N + 1$ sample cases are equal i.e., $\dim\{\mathbf{A}_n\} = \dim\{\mathbf{A}_{OOSP}\} = n$, because the OOSP is within a 95% confidence interval for each predictor features of interest. Then, a map $T: \mathbb{R}^n \to \mathbb{R}^n$ that enforces rigid motion in the low dimensional space is possible if $\|T(\mathbf{A}_n) - T(\mathbf{A}_{OOSP})\| = \|\mathbf{A}_n - \mathbf{A}_{OOSP}\|, \forall \mathbf{A}_n, \mathbf{A}_{OOSP} \in \mathbb{R}^n$, provided that the OOSP does not reside in the tail distribution of each $P$. This constraint is imposed to restrict the dissimilarity matrix, $\mathbf{D}$, from changing significantly. Therefore, Theorem A.1 holds.

**Table A.1** Criteria for evaluating normalized stress and its corresponding interpretation [54].

| 100 x Normalized Stress | Goodness of Fit |
|---|---|
| $\geq 0.2$ | Poor |
| $0.05 < 0.2$ | Good |
| $0 < 0.05$ | Perfect |

# Credit author statement

Ademide O. Mabadeje: Conceptualization, Methodology, Software, Validation, Visualization, Formal analysis, Writing – Original draft.

Michael Pyrcz: Data curation, Conceptualization, Supervision, Funding acquisition, Writing – Reviewing and Editing.

# Declaration of competing interest

The authors declare that they have no known competing financial interests or personal relationships that could have appeared to influence the work reported in this paper.

# Acknowledgements

The authors sincerely appreciate Equinor and the Digital Reservoir Characterization Technology (DIRECT) consortium's industry partners at the Hildebrand Department of Petroleum and Geosystems Engineering, University of Texas at Austin for financial support. We acknowledge Equinor for granting permission to use the Duvernay case study dataset presented here.

# Data Availability

The data and well-documented workflow used is publicly available on the corresponding author's GitHub Repository: https://github.com/Mide478/LowerDimensionalSpace-Stabilization-RT-UQI on publication. No proprietary data is within the GitHub repository and the data used for the synthetic case study is publicly available in the GeoDataSets respository: https://github.com/GeostatsGuy/GeoDataSets/blob/master/unconv_MV_v4.csv.